\documentclass{article}
\usepackage{spconf,amsmath,epsfig}

\let\OLDthebibliography\thebibliography
\renewcommand\thebibliography[1]{
  \OLDthebibliography{#1}
  \setlength{\parskip}{0pt}
  \setlength{\itemsep}{0pt plus 0.3ex}
}

\usepackage{cite}
\usepackage{amsmath,amssymb,amsfonts}
\usepackage{algorithmic}
\usepackage{graphicx}
\usepackage{textcomp}
\usepackage{xcolor}
\usepackage{color}
\usepackage{subfigure}
\usepackage[preprint]{spconfa4}
\let\OLDthebibliography\thebibliography
\renewcommand\thebibliography[1]{
  \OLDthebibliography{#1}
  \setlength{\parskip}{0pt}
  \setlength{\itemsep}{0pt plus 0.3ex}
}

\usepackage{multirow}
\begin{document}\sloppy
\title{Intra-Class Uncertainty Loss Function for Classification}
\name{He Zhu$^{\ast,\ddagger}$, Shan Yu$^{\ast,\dagger,\ddagger}$}
\address{$^\ast$Brainnetome Center, National Laboratory of Pattern Recognition (NLPR),\\Institute of Automation, Chinese Academy of Sciences(CASIA)\\
	$^\dagger$CAS Center for Excellence in Brain Science and Intelligence Technology(CEBSIT)\\
	$^\ddagger$School of Future Technology, University of Chinese Academy of Sciences(UCAS)\\
	 \{he.zhu, shan.yu\}@nlpr.ia.ac.cn}

\maketitle
\begin{abstract}
Most classification models can be considered as the process of matching templates. However, when intra-class uncertainty/variability is not considered, especially for datasets containing unbalanced classes, this may lead to classification errors. To address this issue, we propose a loss function with intra-class uncertainty following Gaussian distribution. Specifically, in our framework, the features extracted by deep networks of each class are characterized by independent Gaussian distribution. The parameters of distribution are learned with a likelihood regularization along with other network parameters. The means of the Gaussian play a similar role as the center anchor in existing methods, and the variance describes the uncertainty of different classes. In addition, similar to the inter-class margin in traditional loss functions, we introduce a margin to intra-class uncertainty to make each cluster more compact and reduce the imbalance of feature distribution from different categories. Based on MNIST, CIFAR, ImageNet, and Long-tailed CIFAR analyses, the proposed approach shows improved classification performance, through learning a better class representation.
\end{abstract}

\begin{keywords}
uncertainty, intra-class margin, image classification, long-tailed classification 
\end{keywords}
\section{Introduction}
\par Image classification for inter-class dispersion and intra-class compactness is a classic topic in computer vision communities such as image recognition \cite{krizhevsky2012imagenet,he2016deep}, object detection \cite{lin2017focal,dai2016r-fcn:,ren2015faster} and face verification \cite{liu2017sphereface,wen2016a}. In recent years, performances of these classification tasks have been largely improved along with the development of deep neural networks(DNN), of which the loss function is a significant component.  
\begin{equation}
    \mathcal L_{CE} = -\frac{1}{N} \sum^N_i log \frac{e^{w_{y_i}^T x_i}}{\sum_{k=0}^K e^{ w_k^T x_i}}
    \label{ce_loss}
\end{equation}
\begin{figure}[h]
\centering
\subfigure[Normal Classification]{
\label{moti_1}
\includegraphics[width=0.48\textwidth]{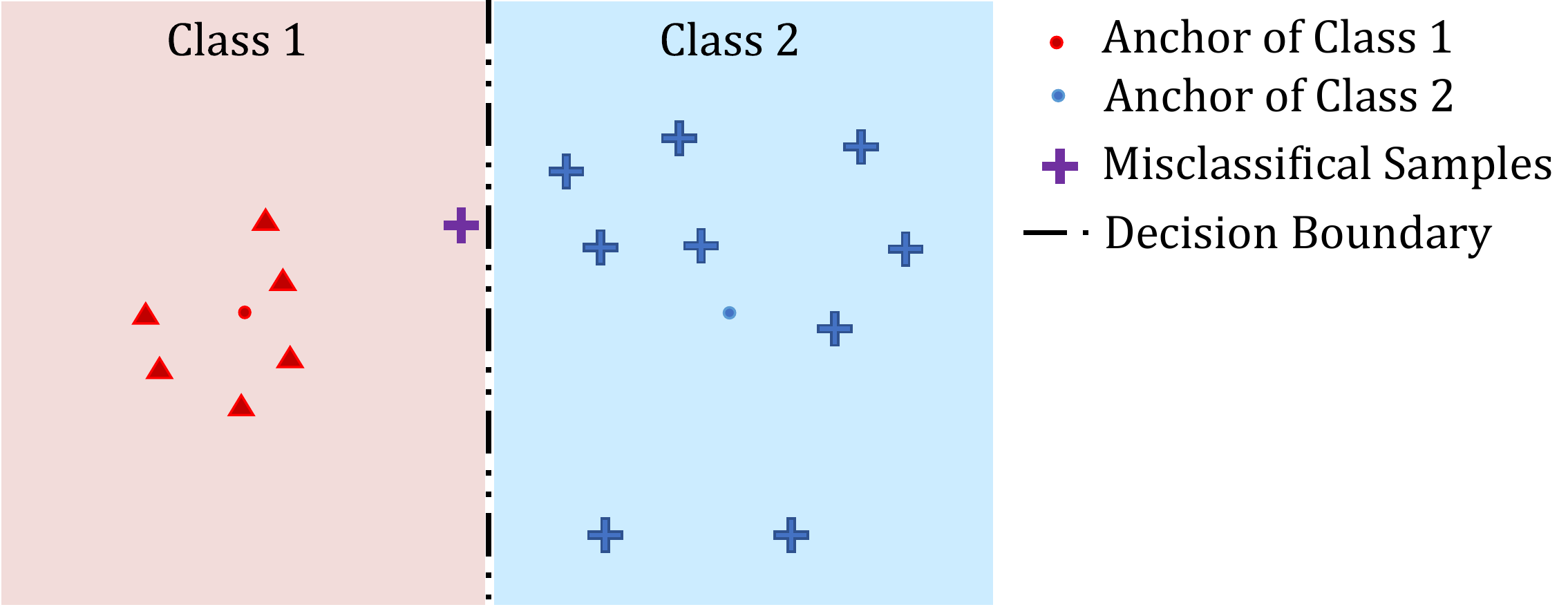}
}
\quad
\subfigure[Consider the uncertainty]{
\label{moti_2}
\includegraphics[width=0.48\textwidth]{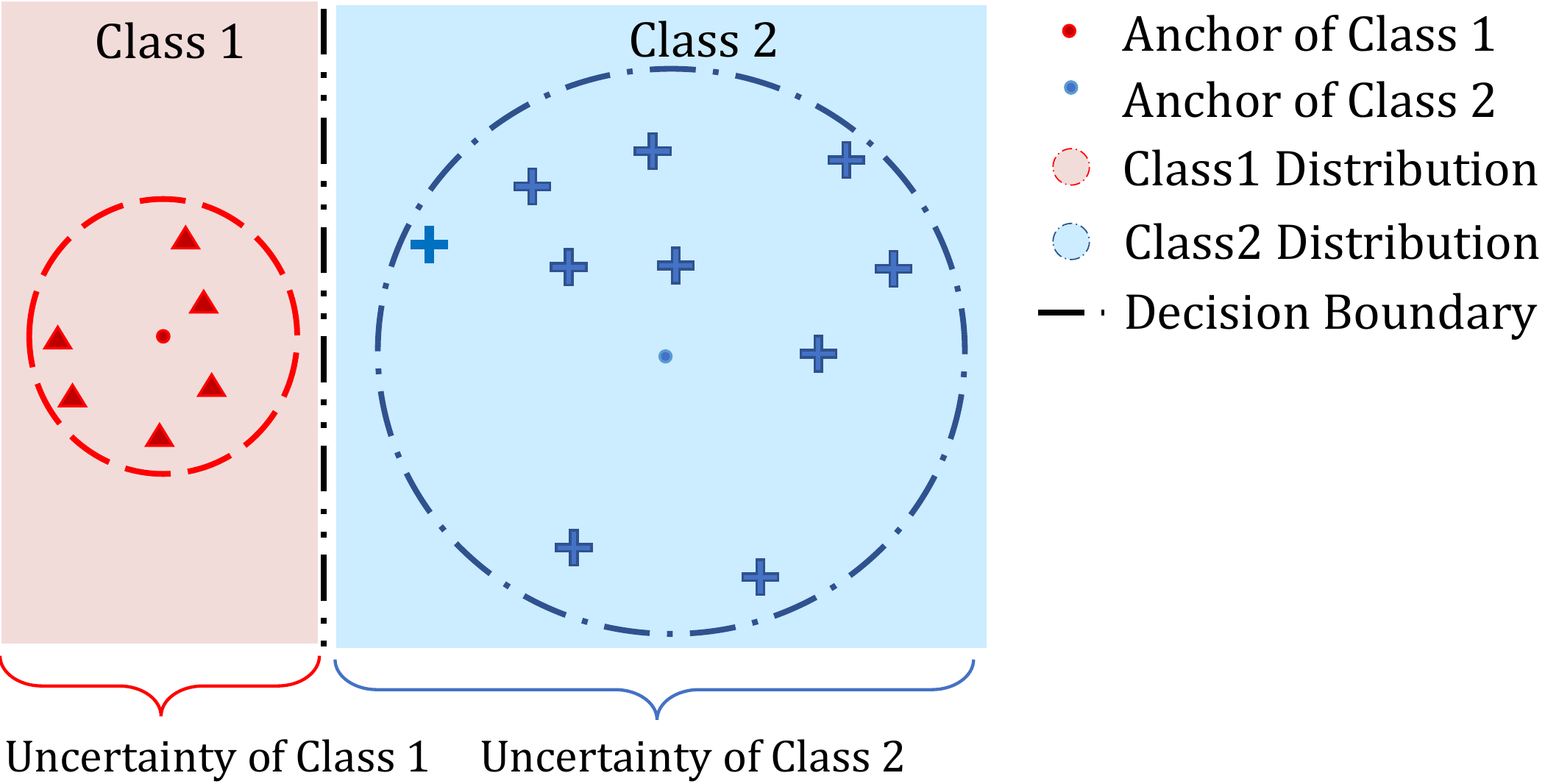}
}
\caption{If the uncertainty of imbalanced categories is not described, the network may make decisions that lead to misclassification because the softmax function only depends on the vector distance between sample features and the anchor point, choosing the minimum one as the output. In this example, because the confounding sample is closer to the anchor of class 1, the purple mark will be classified into the red category. However, if intra-class uncertainty is considered, the sample would more reasonably be classified into the blue category. }
\label{moti}
\end{figure}
\par Softmax loss (Eq.\ref{ce_loss}), which has been widely applied in various tasks, consists of the softmax function and cross entropy and is a distance-based winner-takes-all decision method. Specifically, the anchor of special class k is a row vector of the last linear classifier weight $w_k^T$. This loss function measures the vector distance $w_k^T x_i$ between deep features $x_i$ of samples and anchors of different categories $w_k^T$, which we call decision distance. 
\par
\begin{figure*}[htbp]
\begin{minipage}[t]{0.24\linewidth}
	\centering
	\includegraphics[width=1.0\textwidth, trim=30 20 45 40, clip]{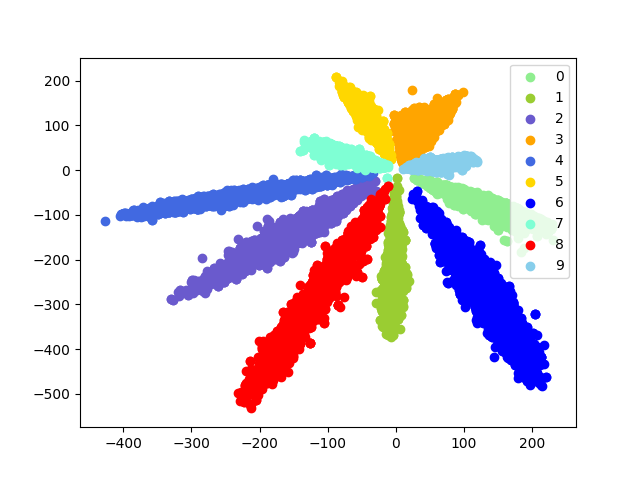}
	\label{fig:s_train}
    \centerline{(a) Softmax loss training}\medskip
\end{minipage} 
\hfill
\begin{minipage}[t]{0.24\linewidth}
	\centering
	\includegraphics[width=1.0\textwidth, trim=40 20 40 40, clip]{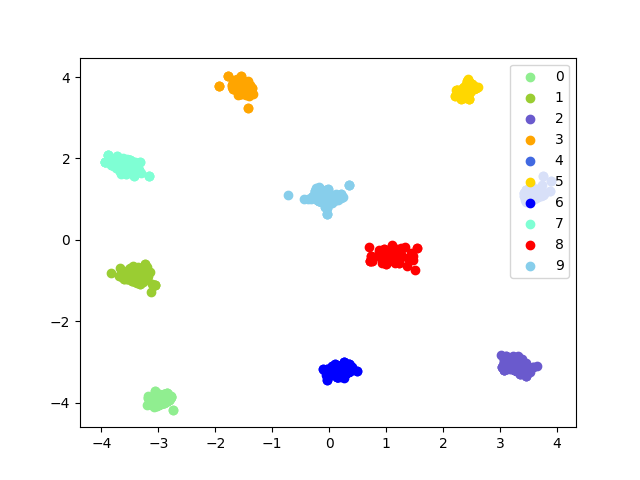}
	\label{fig:c_train}
	\centerline{(b) Center loss training}\medskip
\end{minipage} 
\hfill
\begin{minipage}[t]{0.24\linewidth}
	\centering
	\includegraphics[width=1.0\textwidth, trim=40 20 40 40, clip]{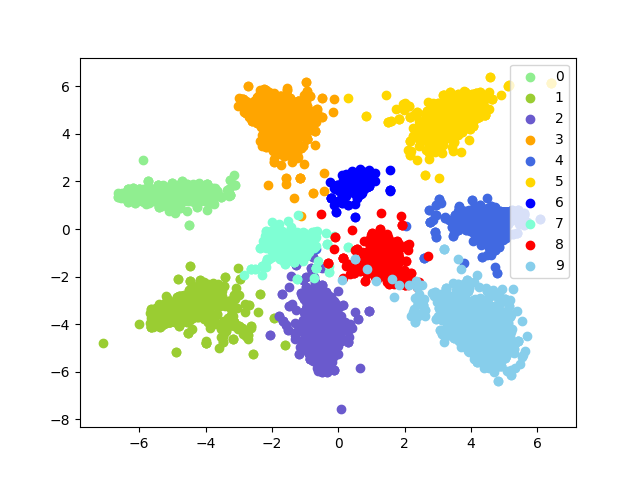}
	\label{fig:l_train}
	\centerline{(c) L-GM loss training}\medskip
\end{minipage} 
\hfill
\begin{minipage}[t]{0.24\linewidth}
	\centering
	\includegraphics[width=1.0\textwidth, trim=40 20 40 40, clip]{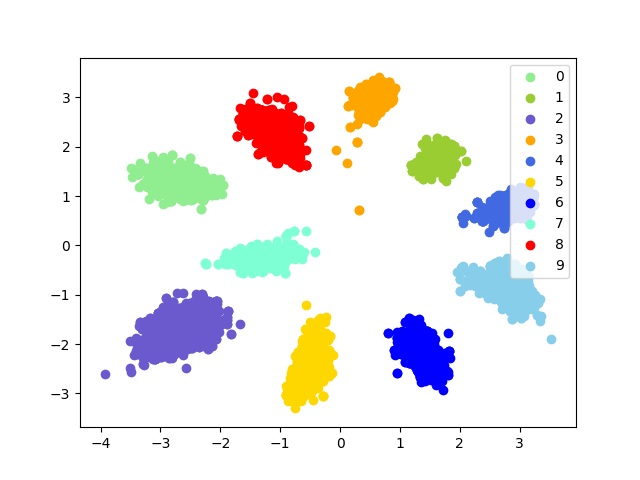}
	\label{fig:o_train}
	\centerline{(d) Our loss training}\medskip
\end{minipage} 
\\
\begin{minipage}[t]{0.24\linewidth}
	\centering
	\includegraphics[width=1.0\textwidth, trim=30 20 45 40, clip]{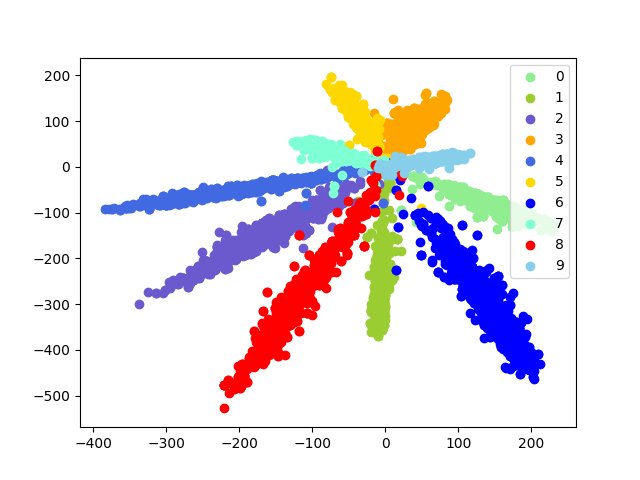}
	\label{fig:s_test}
	\centerline{(e) Softmax loss testing}\medskip
\end{minipage} 
\hfill
\begin{minipage}[t]{0.24\linewidth}
	\centering
	\includegraphics[width=1.0\textwidth, trim=40 20 40 40, clip]{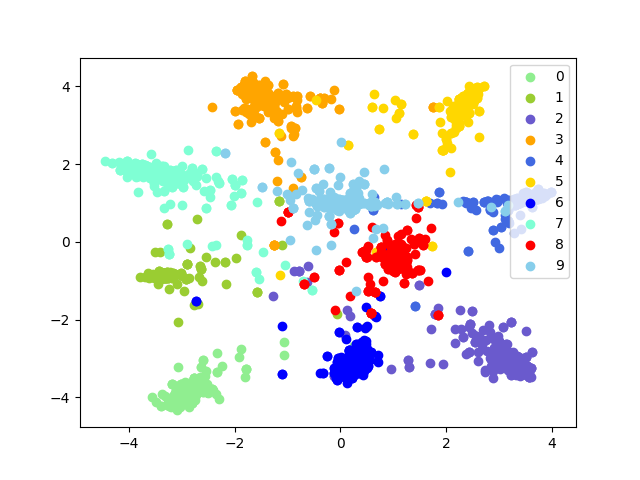}
	\label{fig:c_test}
	\centerline{(f) Center loss testing}\medskip
\end{minipage} 
\hfill
\begin{minipage}[t]{0.24\linewidth}
	\centering
	\includegraphics[width=1.0\textwidth, trim=40 20 40 40, clip]{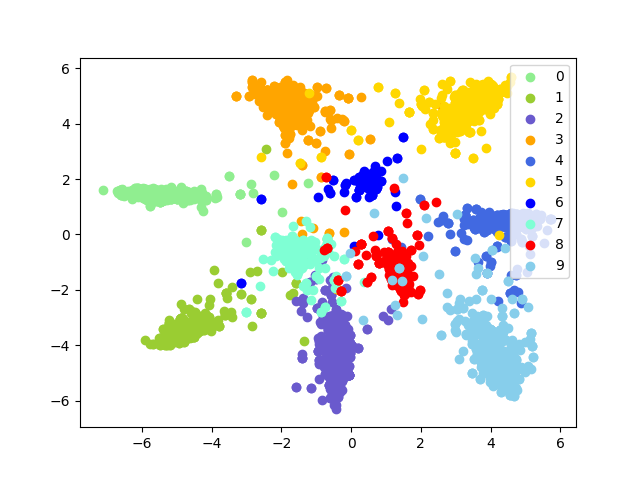}
	\label{fig:l_test}
	\centerline{(g) L-GM loss testing}\medskip
\end{minipage} 
\hfill
\begin{minipage}[t]{0.24\linewidth}
	\centering
	\includegraphics[width=1.0\textwidth, trim=40 20 40 40, clip]{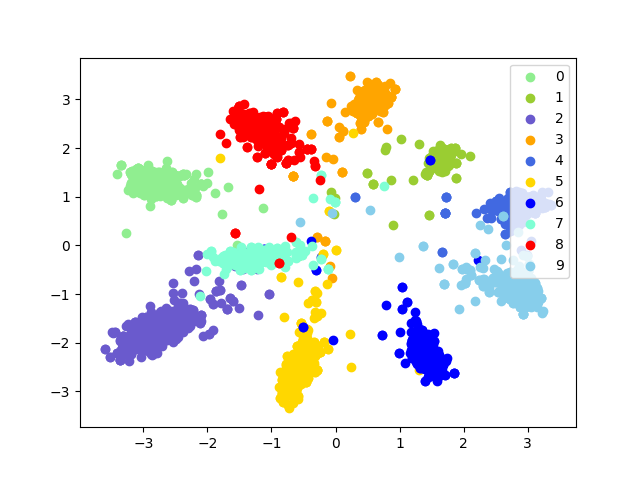}
	\label{fig:o_test}
	\centerline{(h) Our loss testing}\medskip
\end{minipage} 
\caption{ Visualization of MNIST training and testing dataset functions. Even for the simplest MNIST dataset, previous methods experience difficulty in dealing with unbalanced feature distribution, leading to confusion during inference. However, our proposal ensures that samples belonging to the same class are closer to each other, thereby reducing the imbalance in the distribution of features from different classes.}
\label{mnist_v}
\end{figure*}

 Previous loss functions primarily focused on learning a robust anchor point on the training dataset or a better measure of the distance between the feature and anchor point \cite{liu2017sphereface,liu2016largemargin,wang2017normface,wang2018cosface,zheng2018ring,wan2018rethinking,yang2018robust}. However, the anchor is too weak to describe the imbalanced characteristics of different classes of the training dataset, and higher-order information is usually needed to infer the validation dataset. For example, as shown in Fig.\ref{moti}, lack of consideration of the uncertainty of global class distribution will result in network confusion, which can even occur for classification of the simplest MNIST dataset (Fig.\ref{mnist_v} a,e). This issue is difficult to solve by only improving anchor learning or metrics. Our motivation to solve this problem comes from several face verification and detection methods \cite{shi2019probabilistic,choi2019gaussian,he2018bounding}, which model class uncertainty as the variance of the Gaussian Mixed Model (GMM) to learn reliable representation of instances.  
\par In this paper, we propose a method that uses intra-class uncertainty (ICU) as a measure of the imbalance of different classes/categories, and propose ICU loss for image classification, where deep features exhibit independent Gaussian distribution. Our main contributions are as follows:
\begin{itemize}
    \item Compared with previous methods, we introduce ICU as an improved decision distance to calculate the predicted classification labels and learn the parameters describing the prior information of each class from the training dataset automatically.
    \item We propose a novel intra-class margin mechanism, which can learn robust uncertainty representation/information of different classes and reduce the imbalance in the distribution of features from different categories. 
    \item We propose a new regularization method to constrain ICU and overcome the problem of variance parameter convergence.
\end{itemize}
\section{Related work}
\par Softmax loss has been widely used in various tasks and consists of the softmax function and cross entropy. After optimizing softmax loss, the deep features of images extracted by the network tend to follow radial distribution \cite{wang2017normface}. That is, more easily obtained samples are closer to the ground-truth anchor points of each category, and difficult to obtain samples are further away. As shown in Fig.\ref{mnist_v} a,e, this may lead to sample confusion due to rough decision boundaries and unbalanced feature distribution.
\par The main strategy used in studies aimed at designing effective loss function \cite{wang2017normface,wen2016a,zheng2018ring,liu2016largemargin,liu2017sphereface,wang2018cosface} is to make better decisions regarding the distribution of features, due to the fact that the distribution of deep features on the training dataset largely depends on the loss function, as shown in Fig.\ref{mnist_v}. Classification loss functions are usually designed in two forms, i.e., margin-based, and center-based. \\
\noindent \textbf{Margin-based loss}. Margin-based loss functions have a clear geometric interpretation, which tends to increase the distance between classes. Variants redefine various forms of margins, e.g., cosine-margin \cite{liu2016largemargin}, and angular-margin \cite{liu2017sphereface,wang2018cosface}. These proposals can help prevent the network from overfitting and to learn better image representations.
\begin{equation}
    \mathcal L_{Margin} = -\frac{1}{N}\sum^N_i log \frac{e^{-d(w_{y_i},x_i,m)}}{e^{-d(w_{y_i},x_i,m)} + \sum_{k\not=0}^K e^{-d(w_k,x_i)}}
\end{equation}
where $m$ is a margin factor.\\
\noindent \textbf{Center-based loss}. Center-based loss functions \cite{wen2016a,schroff2015facenet,qi2018face,hadsell2006dimensionality,vo2017revisiting} learn anchor points of different categories in feature space. These methods drive the samples to their positive centroid by optimizing center-based regularization. Such co-supervised solutions significantly enhance the discriminative ability of deep learning features.
\begin{equation}
    \mathcal L_{Center} = -\frac{1}{N}\sum^N_i log \frac{e^{w_{y_i}^Tx_i}}{\sum_{k=0}^K e^{w_k^Tx_i}}+\frac{\lambda}{2}\sum_{i}^N ||x_i-c_{y_i}||^2_2
    \label{center}
\end{equation}
\noindent \textbf{L-GM loss}. Previous research \cite{wan2018rethinking} has suggested that use Mahalanobis distance of Gaussian model as the decision distance, with margin and likelihood terms used to promote Gaussian distribution (GD). However, the use of L-GM loss misses an important factor of uncertainty during the inference period, lacking ln$|\Sigma|$ in the decision distance term $d_m$ in Eq.\ref{lgm_loss},\ref{lgm_dis}. Therefore, L-GM cannot solve the problem of class/distribution imbalance.
\begin{equation}
\begin{aligned}
    \mathcal L_{L-GM}=&-\frac{1}{N}\sum_{i=1}^{N} log\frac{e^{-d_{z_i}(1+\alpha)}}{\Sigma _{m}^{M} e^{-d_{m}(1+R(m=z_i)\alpha)}} \\ 
    &+\lambda (d_{z_i}+\frac{1}{2} log|\Lambda_{z_i}|)
\end{aligned}
\label{lgm_loss}
\end{equation}

\begin{equation}
	\begin{aligned}
	    d_{m}=&\frac{1}{2}(x_i-\mu_m)^T\Lambda_m^{-1}(x_i-\mu_m), m\in[1, M]
	\end{aligned}
	\label{lgm_dis}
\end{equation}
\noindent \textbf{Long-tailed Classification}.  As the number of category grows, maintaining a balanced dataset across multiple categories is challenging as data are inherently long-tailed, i.e., with dramatically different number of training samples in individual categories. Therefore, long-tailed classification is the key to deep learning at scale. Recent works \cite{lin2017focal,zhang2017mixup,cui2019class} has started to fill the performance gap between class-balanced datasets and long-tailed datasets, while new long-tailed benchmarks are emerging, such as long-tailed CIFAR-10/-100 for image classification.

\section{Method}
\subsection{Intra-Class Uncertainty Loss}
In the Bayesian view of classification, the network outputs the probabilities of each category, and chooses the maximum value as the inference result. According to the assumption of GMM, the $p(z_i)$ is category prior to the K-classes task. $z_i$ is the ground-truth label of $x_i$, and the posterior probability distribution can be written as:
\begin{equation}
p(z_i|x_i)=\frac{\mathcal N(x_i;\mu_{z_i}, \sigma_{z_i})p(z_i)}{\sum_{k=1}^{K}\mathcal N(x_i;\mu_{k}, \sigma_{k})p(k)}
\label{gmm}
\end{equation}
\begin{equation}
    \mathcal N(x_i;\mu_{k}, \sigma_{k}) = \frac{e^{-\frac{1}{2}(x_i-\mu_k)^T\Sigma_{k}^{-1}(x_i-\mu_k)}}{(2\pi)^{\frac{d}{2}}|\Sigma|^{\frac{1}{2}}}
\end{equation}
where $\mu_k$ and $\sigma_k$ are the parameters of k-th category, and minibatch size is $N$. Assuming that $p(z_i)=p(k)=\frac{1}{K}$, we introduce $p(z_i|x_i)$ into the calculation of cross-entropy and obtain:
\begin{equation}
\begin{aligned}
\mathcal L_{cls}&=-\frac{1}{N}\sum_{i=1}^{N}logp(z_i|x_i)\\
&=-\frac{1}{N}\sum_{i=1}^{N} log\frac{e^{-d_{z_i}}}{\Sigma _{k}^{K} e^{-d_{k}}} + \lambda \mathcal L_{reg}
\end{aligned}
\label{gmmce}
\end{equation}

\begin{equation}
    d_{k}=\frac{1}{2}[(x_i-\mu_k)^T\Sigma_{k}^{-1}(x_i-\mu_k)+ln|\Sigma_{k}|], k\in[1, K] 
    \label{dist}
\end{equation}
\par We propose that $ln|\Sigma_{k}|$ in Eq.\ref{dist} can represent the uncertainty of various categories, and it should be considered as an important factor in decision distance that can affect inference.

\subsection{Regularization $\mathcal L_{reg}$}
\par From the maximum-likelihood of Eq.\ref{gmmce}, the variance parameter is log $d_k$, which is difficult to converge due to ln$| \Sigma|$ in Eq.\ref{dist}. This term is difficult to optimize based on previously suggested solutions \cite{wan2018rethinking,shi2019probabilistic,choi2019gaussian,he2018bounding}, which soften parameter $\Sigma$ as the diagonal matrices \cite{wan2018rethinking} or use a two-step learning strategy \cite{he2018bounding}. However, these methods disregard ICU information and can cause the problems observed in Fig.\ref{moti}. Parameters $\Sigma$ is challenging to learn when applying maximum-likelihood estimation.In contrast, a solution based on the theory of statistics and moment estimation can effectively optimize $\Sigma$. From this perspective, we propose ICU loss $\mathcal L_{ICU}$. 
\par First, considering the partial derivatives of the two distribution parameters, $\mu$ and $\sigma$ should be 0 when parameters converge:
\begin{equation}
    \frac{\partial ln\ {p(x_i;\mu_k,\sigma_k)}}{\partial \mu_k} = 0
\end{equation}
\begin{equation}
    \frac{\partial ln\ { p(x_i;\mu_k,\sigma_k)}}{\partial \sigma_k} = 0
\end{equation}
Further, we have:
\begin{equation}
    \mu_k=\frac{1}{N}\sum_{i=1}^{N}x_{z_i=k}=\bar{\mu}_{N_k}
\end{equation}	
\begin{equation}
    \sigma_k^2= \frac{1}{N}\sum_{i=1}^{N} (x_{i}-\mu_k)^T(x_{i}-\mu_k)=\bar {\sigma}_{N_k}^2
    \label{sig}
\end{equation}
where $\bar{\mu}_N$ and $\bar {\sigma}_{N}^2$ are average results of mini-batch size samples.  Regularization $\lambda \mathcal L_{reg}$ aims to prevent the non-convergence of using maximum-likelihood to learn the parameters.
\begin{equation}
\begin{split}
\lambda \mathcal L_{reg}=&\sum_{k=1}^{K}\lambda_1| \mu_k-\bar{\mu}_{N_k}|^2+\lambda_2|\sigma_k^2-\bar {\sigma}_{N_k}^2|^2
\end{split}
\label{reg}
\end{equation}
\par $|\mu_k-\bar{\mu}_{N_k}|^2$ in Eq.\ref{reg} has the same meaning as the Eq.\ref{center} in center loss. Here, we simply prove the gradient of parameters $\mu_k$ and $\sigma_k$.
\par According to Eq.\ref{sig} and Eq.\ref{reg}:
	\begin{equation}
   	\begin{split}
    \frac{\partial L_{reg}}{\mu_k}=[ 2\eta_1 + 4\eta_2(\sigma_k^2-\sigma_{N_k}^2)](\mu_k-\bar{\mu}_{N_k})
\end{split}
\label{grad}
\end{equation}
\begin{equation}
\begin{split}
    \frac{\partial L_{reg}}{\sigma_k}=4\eta_2(\sigma_k^2-\bar{\sigma}_{N_k}^2)\sigma_k\\
\end{split}
\end{equation}
\par From Eq.\ref{grad}, the gradient of $\mu_k$  is different from the center loss in Eq.\ref{center}, and the ICU loss softens the cluster constraint on the centroid. In Fig.\ref{mnist_v} b,f, the center loss tends to overfit the crowded population of the training dataset and performs poorly on the testing dataset. Our proposal removes the constraints and tends to be consistent with the training and testing datasets, as shown in Fig.\ref{mnist_v} d,h.
\par Compared with L-GM loss, ICU is applied to the Mahalanobis distance for more accurate measurement. In addition, we propose Eq.\ref{reg} to learn the variance efficiently and avoid the non-convergence of optimizing ln$|\Sigma|$.
\par Moreover, we introduce two margins: i.e., inter-class margin and intra-class margin, into Eq.\ref{icu} and Eq.\ref{dzi} for better clustering. The large margin loss function can be formulated as:
\begin{equation}
\mathcal L_{ICU}=-\frac{1}{N}\sum_{i=1}^{N} log\frac{e^{-d_{z_i}(1+\alpha)}}{\Sigma _{k,k \not = z_i}^{K} e^{-d_{k}}+e^{-d_{z_i}(1+\alpha)}}
\label{icu}
\end{equation}
where,
\begin{equation}
	\begin{aligned}
		d_{z_i}=\frac{1}{2}[(x_i-\mu_{z_i})^T\Sigma_{{z_i}}^{-1}(x_i-\mu_{z_i})\\&+ln(1+\gamma)|\Sigma_{z_i}|]
	\end{aligned}
\label{dzi}
\end{equation}
\begin{equation}
	\begin{aligned}
		d_{k}=\frac{1}{2}[(x_i-\mu_k)^T\Sigma_{k}^{-1}(x_i-\mu_k)\\&+ln|\Sigma_{k}|], k\in[1, K]
	\end{aligned}
\label{dk}
\end{equation}
where $\alpha$ is the hyperparameter for creating a large margin between different categories, and $\gamma$ can adjust the intra-class margin.\\	

\subsection{\textbf{Margin Mechanism}}
\begin{figure}[h]
\centering
\subfigure[A geometry interpretation of the inter-class and intra-class margin during training samples from class 1]{
\label{class_1}
\includegraphics[width=0.47\textwidth]{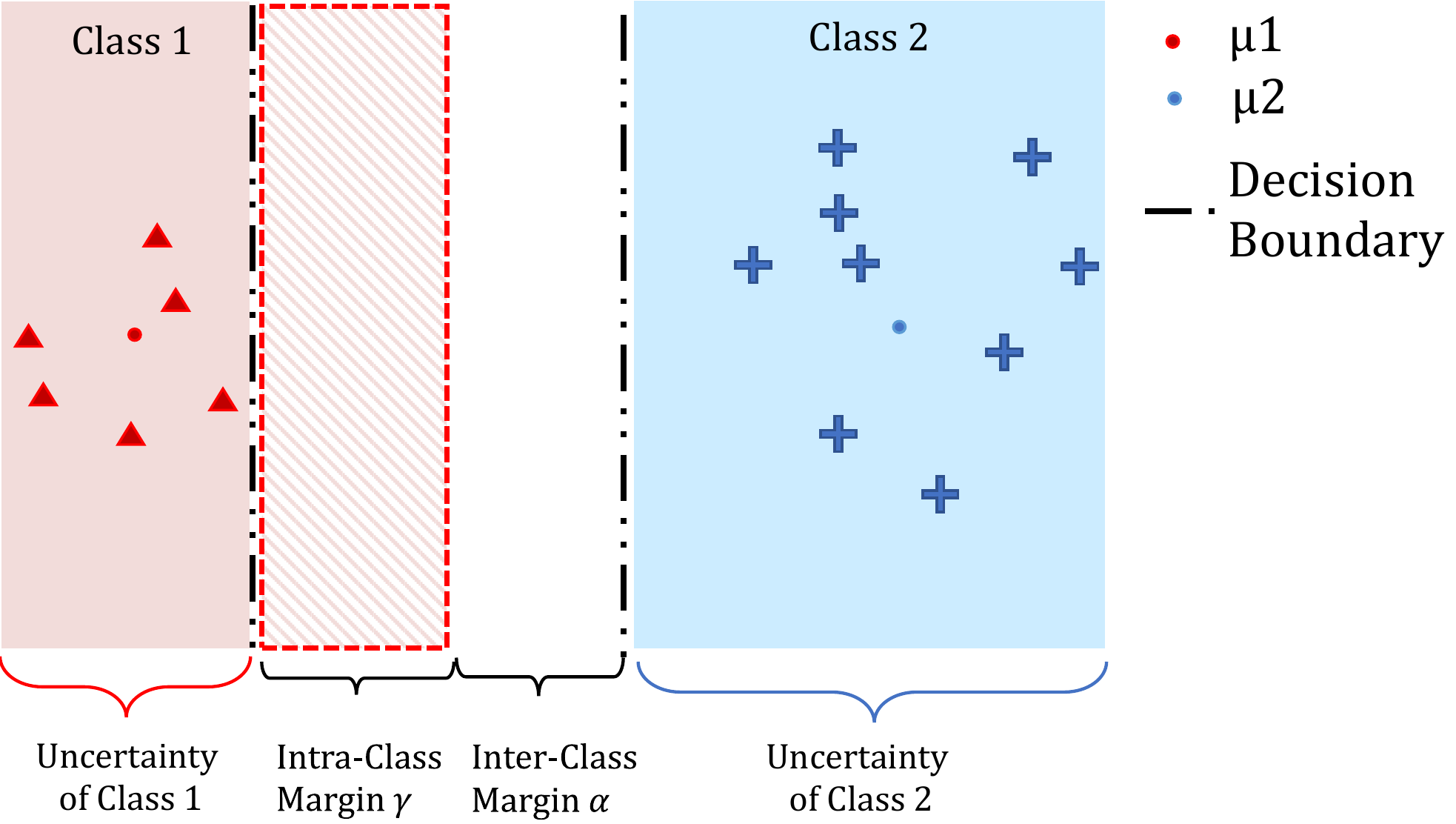}
}
\quad
\subfigure[A geometry interpretation of the inter-class and intra-class margin during training samples from class 2]{
\label{class_2}
\includegraphics[width=0.47\textwidth]{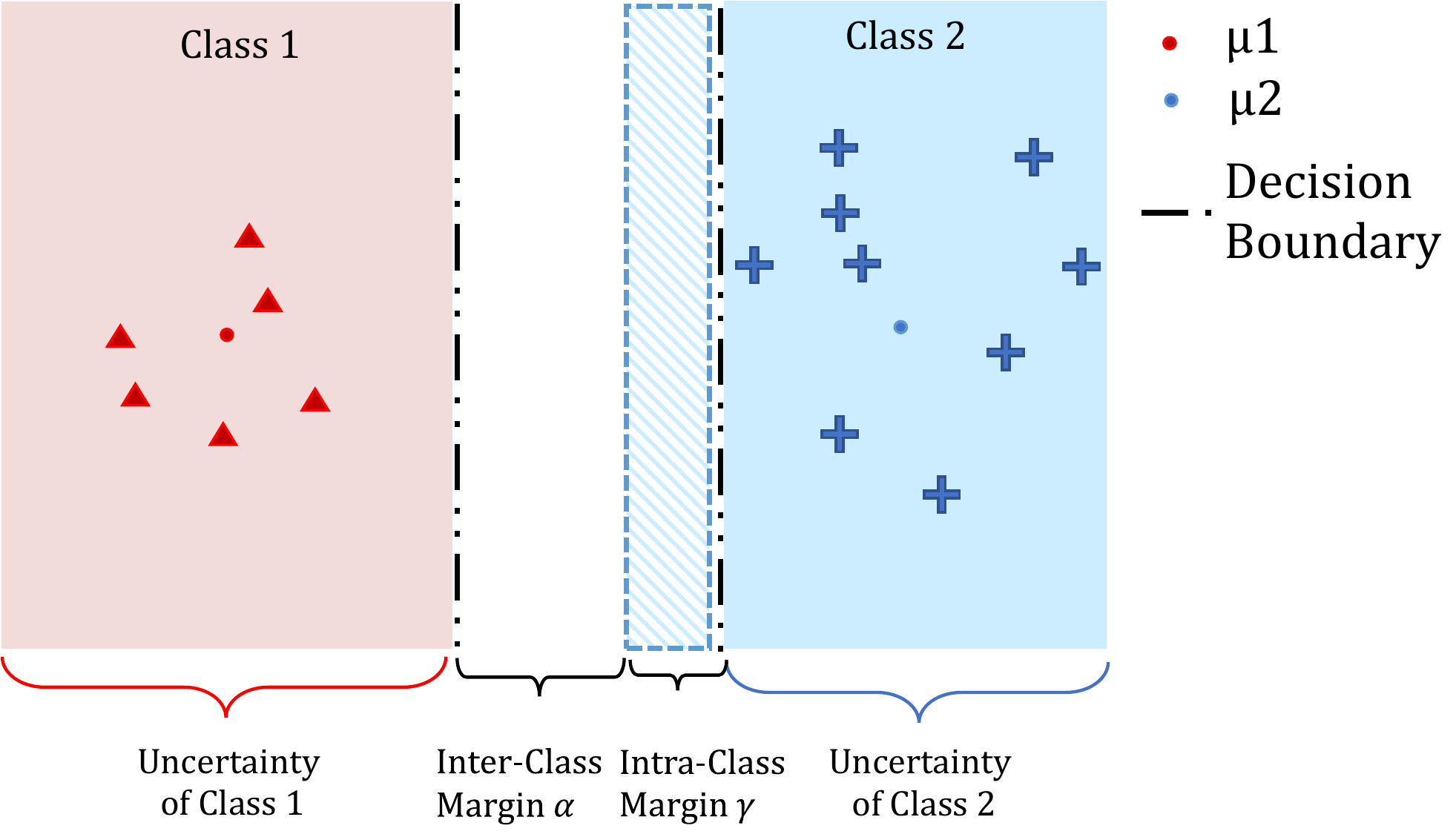}
}
\caption{Parameters $\alpha$ and $\gamma$ add inter-class and intra-class margins to the decision boundary between two categories. $\alpha$ is used to ensure the distances between the deep features of $x_i$ from different classes are far from each other in order to learn more robust anchors; $\gamma$ reduces the uncertainty of the ground truth class of $x_i$ to learn a robust uncertainty representation. }
\label{fig:ab}
\end{figure}
The margin mechanism of loss function is a commonly used technique to improve network generalization performance \cite{liu2016largemargin,liu2017sphereface,wang2018cosface,HuCY19}. In this work, we use parameters $\alpha, \gamma$ to adjust the margin, where $\alpha$ is the widely used margin between different classes. Importantly, we present $\gamma$ as the intra-class margin based on the uncertainties of various categories. From the perspective of feature distribution, the schematic is shown in Fig.\ref{fig:ab}. Note, the $\alpha$ and $\gamma$ parameters are set to 0 during testing.

\par Here, $\alpha$ is the mechanism to make $x_i$ closer to the feature mean of class $z_i$ than the feature mean of other classes, and promote $x_i$ to belong to its ground truth class $z_i$, it should satisfy that: 
\begin{equation}
    e^{-(1+\alpha)d_{z_i}} > e^{-d_k} \Longleftrightarrow d_k-d_{z_i} > \alpha d_{z_i}, \forall k \not = z_i
\end{equation}
\begin{table*}[tbp]
    \caption{Classification performance on various standard classification datasets.}
    \centering
    \begin{tabular}{c|c|c|c|c|c c}
    \hline
    \hline
    Loss Functions & MNIST 2-D & MNIST 100-D & CIFAR-10 & CIFAR-100 & ImageNet Top1 & ImageNet Top5 \\
    \hline
    Networks & \multicolumn{2}{c|}{6-layer CNN} & ResNet-18 & VGG-13 & \multicolumn{2}{c}{ResNet-101} \\
    \hline
    Baseline     &  98.18 $\pm$ 0.01 & 99.32 $\pm$ 0.01 & 93.41 $\pm$ 0.05 & 74.39 $\pm$ 0.07  & 76.50 $\pm$ 0.2& 92.45$\pm$0.08\\
    Center \cite{wen2016a}    & 98.55 $\pm$ 0.01& 99.53 $\pm$ 0.01 & 94.06 $\pm$ 0.02 & 75.15 $\pm$ 0.06  & 76.55 $\pm$ 0.2& 92.79$\pm$0.08 \\
    L-Softmax \cite{liu2016largemargin} & 98.70 $\pm$ 0.02&99.57 $\pm$ 0.01  & 93.95 $\pm$ 0.04 & 75.17 $\pm$ 0.05 & - &-\\
    L-GM \cite{wan2018rethinking} & 98.83 $\pm$ 0.01& 99.61 $\pm$ 0.01 & 93.99 $\pm$ 0.05 & 76.06 $\pm$ 0.08  & 76.65 $\pm$ 0.2& 92.84 $\pm$0.08 \\
    \hline
    \textbf{ICU} & \textbf{99.04 $\pm$ 0.01}&\textbf{99.66 $\pm$ 0.01}  & \textbf{94.38$\pm$ 0.05} & \textbf{77.80 $\pm$ 0.08}  & \textbf{77.10 $\pm$ 0.2}&\textbf{93.29 $\pm$ 0.08} \\
    \hline
    \hline
    \end{tabular}
    \label{tab:classification}
\end{table*}
\par To understand the role of $\gamma$ in ICU loss, a simpler case may be considered, where $\alpha=0$, and $x_i$ is in the centroid of two classes, i.e. $x_i$ has the same Mahalanobis distance to feature mean of $z_i$ and $k$ classes, which is easy to confuse when using L-GM loss or other methods, as shown in Fig.\ref{moti}. Then, if Eq.\ref{bet} holds true, $x_i$ is classified to class $z_i$, indicating that the uncertainty of $x_i$ belonging to class $z_i$ should be smaller than that to other classes. 
\begin{equation}
e^{-d_{z_i,\gamma}} > e^{-d_k} \Longleftrightarrow{\sigma^2_{k} - \sigma^2_{z_i}} >  \gamma \sigma^2_{z_i}, \forall k \not = z_i
\label{bet}
\end{equation}

\par Previous methods \cite{wang2018additive,wang2018cosface,wan2018rethinking} define the distance and margin in the feature space, and mainly used geometric distance between the feature and anchor point. However, our definition of distance includes ICU because prior information can be biased against inference. This adaptability takes into account the divergence distribution of different classes.

\section{Experiment}
\subsection{Implement details}
\par For all classification experiments, we set $\lambda_{\mu}$=1e-1,$\lambda_{\sigma}$=1e-1,$\alpha$=1e-4,$\gamma$=1e-3 as constants. 
\par In the standard classification task, we used different settings to test our loss. In the MNIST experiments, we used a two-dimensional (2-D) network consisting of four convolution layers and two fully connected (FC) layers, and a 100-D network consisting of five convolution layers and a FC layer. Batch normalization was applied. During training, we used the Adam optimizer with an initial learning rate of 0.01, weight decay of 0.001, and batch size of 128. In the CIFAR-10 experiments, we used the standard ResNet structures \cite{he2016deep} and Adam optimizer, with a learning rate of 0.01, weight decay of 0.001, and batch size of 256. In the CIFAR-100 experiments, we used the same network structure and settings as in previous research \cite{wan2018rethinking} to compare fairly by only change the loss function. The ImageNet experiment was based on the ILSVRC2012 dataset, which uses the ResNet-101 network on eight 1080ti GPUs for 100 epochs. The batch size was 256 and the initial learning rate was 0.01, which was reduced to 0.001 in the 50th and 0.0001 in the 75th epoch. Dropout was not used, and the weight decay was 0.0005.
\par The long-tailed dataset experiments were based on an open-source baseline of long-tailed classification $\footnote{https://github.com/KaihuaTang/Long-Tailed-Recognition.pytorch/}$.

\subsection{Classification on Standard Datasets}
Table.\ref{tab:classification} shows the classification performance of ICU loss on various standard datasets, compared with other loss functions. The accuracy benchmark of the MNIST dataset is 98.8\% with 2-D FC and 99.6\% with 100-D FC when using an additional PReLU layer. Here, performance between the ICU and other loss functions was compared using the same network structure. In both the CIFAR and ImageNet experiments, ICU loss led to a more accurate result compared with other loss functions. 
\subsection{Classification on the Long-tailed Datasets}
\begin{table}[h]
	\caption{Top-1 accuracy on long-tailed CIFAR-10/100 with different imbalance ratios. All models used the same ResNet-32 backbone.}
	\begin{center}
		\begin{tabular}{c|c|c|c|c|c|c}
		    \hline
			\hline
			{Methods} & \multicolumn{3}{c|}{CIFAR-100-LT}& \multicolumn{3}{c}{CIFAR-10-LT}\\
			\hline
			 Radio & 100 & 50 & 10 & 100 & 50 & 10\\
			\hline
			\hline
			L-GM \cite{wan2018rethinking} &  37.3 & 41.7 & 57.9 & 72.0 & 77.1 & 87.3\\
			Focal loss \cite{lin2017focal} & 38.4& 44.3 & 55.8 & 70.4 & 76.7 & 86.7 \\
			Mixup \cite{zhang2017mixup} & 39.5& 45.0 & 58.0 & 73.1 & 77.8& 87.1\\
			CB Loss\cite{cui2019class}  & 39.6 & 45.2 & 58.0 & \textbf{74.6} & \textbf{79.3} & 87.1\\
			\hline
			\textbf{ICU} & \textbf{39.8}&\textbf{45.4} & \textbf{58.8} & 73.4 & 79.2 & \textbf{87.7}\\
			\hline
			\hline
		\end{tabular}
	\end{center}
	\label{longtail}
\end{table}

\begin{table}[htbp]
\caption{Results of ICU loss variants in top-1 accuracy (\%) on CIFAR-100/LT-CIFAR-100 (Radio10) datasets.}
    \centering
    \begin{tabular}{c|c|c|c|c}
        \hline
        \hline
        Module &  $\alpha $ &  $\gamma$ & CIFAR-100 & LT-CIFAR-100\\
        \hline
        Baseline & &  &  74.39  & 55.8 \\
        ICU loss & & &  76.52  &  58.0 \\
        - &\checkmark &  &  77.06   &  58.2 \\
        - & & \checkmark &  77.37   &  58.4 \\
        - & \checkmark & \checkmark &  \textbf{77.80} & \textbf{58.8}\\
        \hline
        \hline
    \end{tabular}
    \label{ablation}
\end{table}
\par Long-tail datasets are suitable for verifying the benefits of ICU loss in solving the imbalance problem, which have significantly different numbers of training instances of different classes. We compared ICU loss with previous methods in long-tailed classification: i.e., hard example mining \cite{lin2017focal} and the re-balancing training strategies \cite{zhang2017mixup,cui2019class}. Experiments showed that ICU loss improved the classification performance of networks on unbalanced CIFAR-10/100 datasets (Table.\ref{longtail}).

\subsection{Ablation Study}
\par We presented ablations on ICU loss to visualize its behavior and performance. Table.\ref{ablation} shows the results of ICU loss variants with different margin mechanisms. We found that the two margins introduced here, namely $\alpha$ and $\gamma$, can each improve the performance when implemented alone. Importantly, they can work together synergistically to yield the best results.

\section{Conclusions}
\par In this paper, we propose a novel ICU loss function for image classification. The ICU loss introduces uncertainty of various categories to make predictions. Experiments on several classification benchmarks indicated the effectiveness of our proposed approach.

\bibliographystyle{IEEEbib}
\bibliography{root}

\end{document}